\documentclass[sigconf, screen, nonacm]{acmart}
\AtBeginDocument{%
  }

\usepackage{bm}
\usepackage{fontawesome5}
\usepackage{balance}
\begin{document}

\title{Tora3: Trajectory-Guided Audio-Video Generation with Physical Coherence}


\author{Junchao Liao}
\orcid{0000-0003-4282-0843}
\affiliation{%
  \institution{Alibaba Group}
    \city{Hangzhou}
    \country{China}
}
\email{liaojunchao.ljc@alibaba-inc.com}

\author{Zhenghao Zhang}
\authornote{Corresponding author.}
\orcid{0009-0006-7229-1398}
\affiliation{%
  \institution{Alibaba Group}
    \city{Hangzhou}
    \country{China}
}
\email{zhangzhenghao.zzh@alibaba-inc.com}

\author{Xiangyu Meng}
\orcid{0009-0001-6224-7979}
\affiliation{%
  \institution{Alibaba Group}
    \city{Hangzhou}
    \country{China}
}
\email{xulei.mxy@alibaba-inc.com}

\author{Litao Li}
\orcid{0009-0001-0874-9679}
\affiliation{%
  \institution{Alibaba Group}
    \city{Hangzhou}
    \country{China}
}
\email{lilitao.llt@alibaba-inc.com}

\author{Ziying Zhang}
\orcid{0009-0007-1706-1669}
\affiliation{%
  \institution{Alibaba Group}
    \city{Hangzhou}
    \country{China}
}
\email{jiala.zzy@alibaba-inc.com}

\author{Siyu Zhu}
\orcid{0009-0004-5817-6831}
\affiliation{%
  \institution{Fudan University}
    \city{Shanghai}
    \country{China}
}
\email{siyuzhu@fudan.edu.cn}

\author{Long Qin}
\orcid{0009-0003-4712-8570}
\affiliation{%
  \institution{Alibaba Group}
    \city{Hangzhou}
    \country{China}
}
\email{ql362507@alibaba-inc.com}

\author{Weizhi Wang}
\orcid{0009-0003-7874-9468}
\affiliation{%
  \institution{Alibaba Group}
    \city{Hangzhou}
    \country{China}
}
\email{wangweizhi.wwz@alibaba-inc.com}


\renewcommand{\shortauthors}{Junchao Liao et al.}

\begin{abstract}
Audio-video (AV) generation has recently made strong progress in perceptual quality and multimodal coherence, yet generating content with plausible motion-sound relations remains challenging. Existing methods often produce object motions that are visually unstable and sounds that are only loosely aligned with salient motion or contact events, largely because they lack an explicit motion-aware structure shared by video and audio generation. We present Tora3, a trajectory-guided AV generation framework that improves physical coherence by using object trajectories as a shared kinematic prior. Rather than treating trajectories as a video-only control signal, Tora3 uses them to jointly guide visual motion and acoustic events. Specifically, we design a trajectory-aligned motion representation for video, a kinematic-audio alignment module driven by trajectory-derived second-order kinematic states, and a hybrid flow matching scheme that preserves trajectory fidelity in trajectory-conditioned regions while maintaining local coherence elsewhere. We further curate PAV, a large-scale AV dataset emphasizing motion-relevant patterns with automatically extracted motion annotations. Extensive experiments show that Tora3 improves motion realism, motion-sound synchronization, and overall AV generation quality over strong open-source baselines. Project page: \url{https://ali-videoai.github.io/tora3_page}.
\end{abstract}





\maketitle

\section{Introduction}
\label{sec:intro}

\begin{figure*}[!t]
  \centering
  \includegraphics[width=0.8\textwidth]{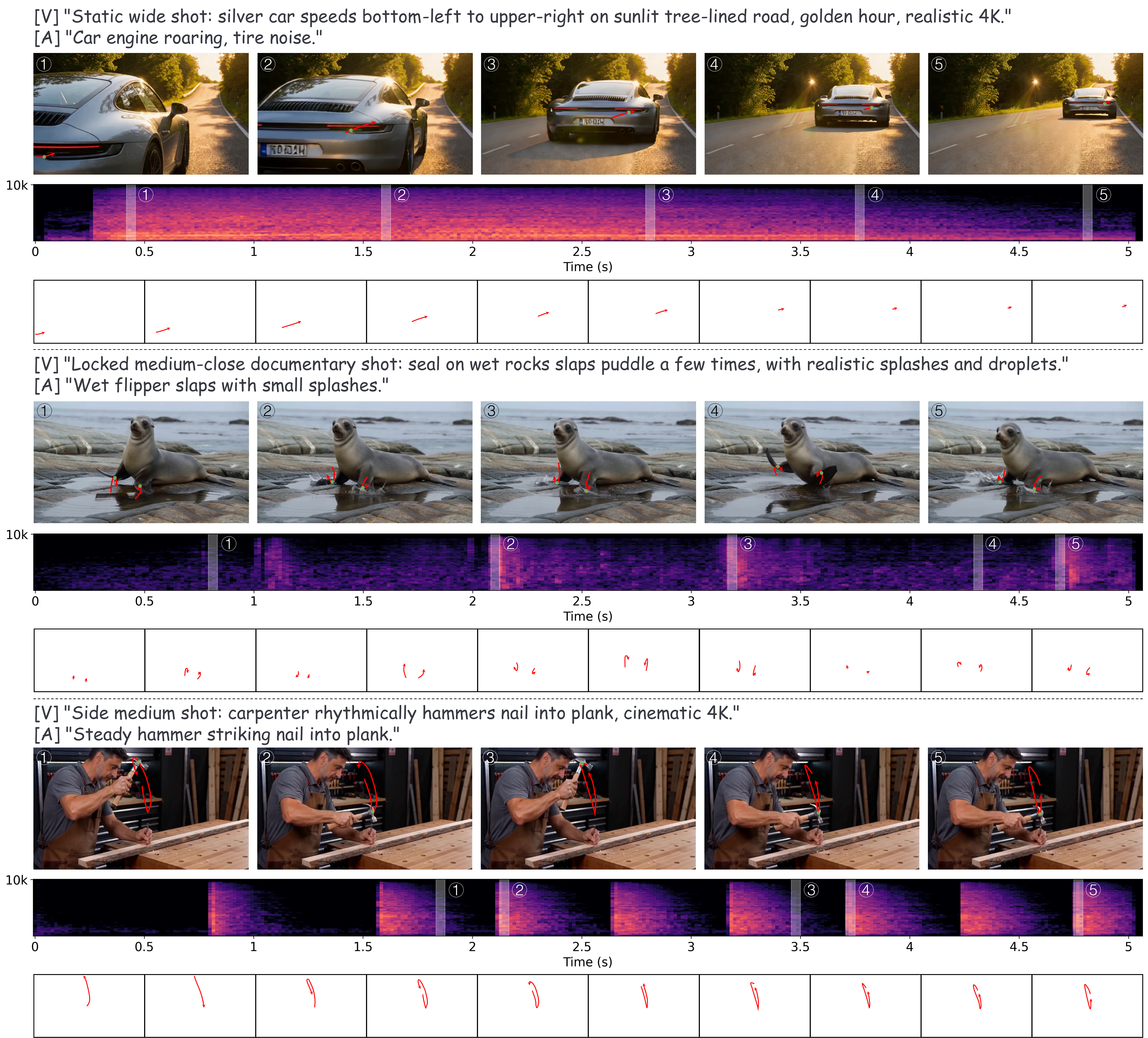}
\caption{Examples generated by Tora3. Tora3 generates audio-video content that follows prescribed trajectories with improved physical coherence, producing more plausible motion and better motion-sound alignment. Long trajectories are segmented for clarity.}
  \Description{A gallery of example results generated by Tora3, showing video frames of objects moving along prescribed trajectories together with their aligned audio, illustrating physically coherent motion and motion-sound alignment.}
  \label{f:showcase}
\end{figure*}

Recent advances in generative modeling have greatly improved video synthesis across modalities, including text-to-video (T2V)~\cite{wan2025wan, blattmann2023stable, kong2024hunyuanvideo, zhang2025waver}, audio-to-video (A2V)~\cite{cheng2025mmaudio, shan2025hunyuanvideo, wang2025kling}, and video-to-audio (V2A)~\cite{chen2025hunyuanvideo, gao2025wan, gan2025omniavatar}. More recently, unified text-to-audio-video (T2AV) models~\cite{hacohen2026ltx2efficientjointaudiovisual, low2025ovi, sora2, wan2.5, veo3} jointly synthesize visual and acoustic content from a single prompt. Commercial systems such as Veo3~\cite{veo3}, Sora2~\cite{sora2}, and Wan2.5~\cite{wan2.5} suggest that high-fidelity multimodal generation is becoming increasingly practical.

Despite this progress, current T2AV generation is still driven primarily by semantic alignment rather than explicit motion-sound modeling. As a result, even strong open-source models~\cite{hacohen2026ltx2efficientjointaudiovisual, low2025ovi, wang2025universe} often produce visually implausible motion or audio only weakly synchronized with salient physical events~\cite{kang2024far}. Visually, generated objects may follow inconsistent trajectories, fall implausibly, or accelerate abruptly, breaking temporal continuity~\cite{deng2025denoising, chefer2025videojam}. Acoustically, sound is often detached from motion: impact sounds may miss the moment of contact, intensity may not reflect interaction strength, and continuous sounds such as sliding or driving may not evolve with motion. Existing methods thus lack an explicit intermediate representation that constrains both modalities under a shared motion-aware structure.

In this work, we focus on a practical notion of \textbf{physical coherence} in audio-video generation: whether object motion follows plausible trajectory-level dynamics, whether salient sound events are temporally aligned with motion or contact, and whether audio intensity evolves with motion strength. We argue that object trajectories offer a natural shared representation for introducing such structure: they compactly encode kinematic cues such as event timing, motion intensity, and temporal evolution, constraining plausible motion in video while indicating when sounds should occur and how their amplitude should evolve. Although not fully specifying real-world physics, trajectories provide a lightweight, interpretable prior for motion-sound coherence. Figure~\ref{f:no_traj} shows that trajectory guidance better aligns visual motion with corresponding audio events.

\DeclareRobustCommand{\mathcircled}[1]{%
  \ooalign{\hfil$#1$\hfil\crcr$\bigcirc$}%
}
\begin{figure}[!t]
    \centering
    \includegraphics[width=0.45\textwidth]{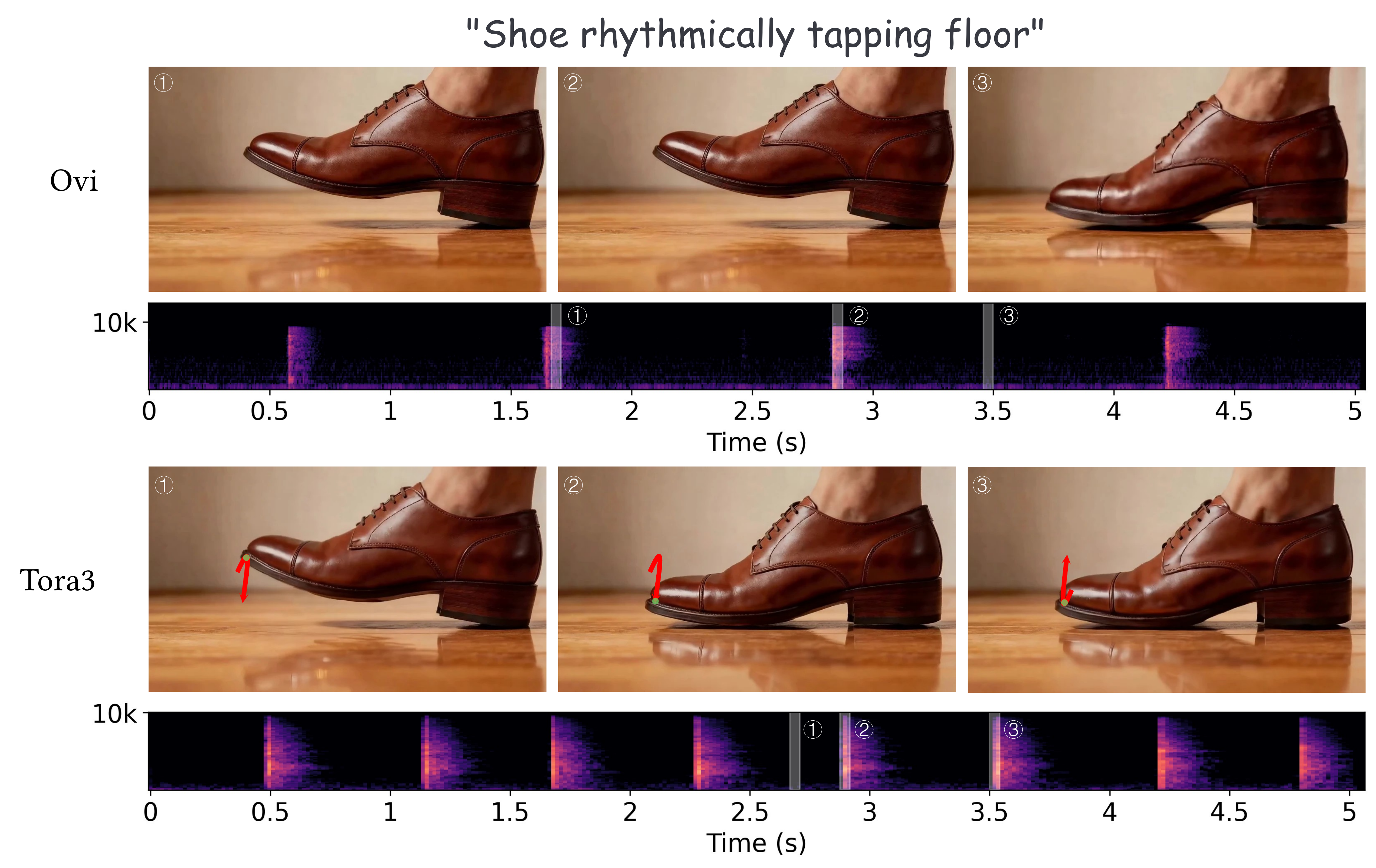}
    \caption{Effect of trajectory guidance on audio-video synchronization. Tora3 better aligns footstep sounds with shoe-ground contact. The red arrow indicates the trajectory from 2.5 to 3.0\,s ($\mathcircled{1}$, $\mathcircled{2}$) and from 3.5 to 4.0\,s ($\mathcircled{3}$).}
    \Description{Comparison of audio-video synchronization with and without trajectory guidance, where red arrows mark the walking trajectory across time intervals and show that footstep sounds are better aligned with shoe-ground contact when trajectory guidance is used.}
    \label{f:no_traj}   
\end{figure}

We present Tora3, a trajectory-guided audio-video generation framework that improves physical coherence through shared kinematic conditioning (see Figure~\ref{f:showcase}). Rather than using trajectories as an auxiliary control signal, Tora3 treats them as a shared kinematic prior to jointly guide video motion and audio generation, with three key components. First, for the video branch, we propose an efficient \textit{trajectory-aligned motion representation} that reuses first-frame latents along sparse trajectories to inject motion cues directly into the latent space without an explicit motion encoder, avoiding distribution shift while preserving motion conditioning. Second, for the audio branch, we design a \textit{kinematic-audio alignment} module. Inspired by second-order motion representations~\cite{liu2024physgen, yuan2025newtongen}, we derive compact kinematic descriptors, including position, velocity, acceleration, and their magnitudes, and inject them into the audio diffusion model via cross-attention, providing direct cues for event timing and motion intensity so that loudness and temporal evolution follow the underlying motion. Third, we introduce a \textit{hybrid flow matching} mechanism with separate probability flows for trajectory and background regions, preserving faithful trajectory-guided motion during early denoising while maintaining local appearance coherence during later refinement. We further curate PAV, a large-scale audio-video dataset of 460k clips emphasizing motion-relevant patterns with automatically extracted motion annotations for trajectory- and kinematics-aware AV generation.

Our main contributions are as follows:

\begin{itemize}
\item We propose \textbf{Tora3}, a trajectory-guided audio-video generation framework that uses object trajectories as a shared kinematic prior to jointly guide visual motion and acoustic events, improving physical coherence in generated content.

\item We develop a unified trajectory-grounded generation framework with three key components: (1) a \emph{trajectory-aligned motion representation} for direct video latent conditioning, (2) a \emph{kinematic-audio alignment} module based on trajectory-derived position, velocity, and acceleration cues, and (3) a \emph{hybrid flow matching} scheme that preserves motion fidelity while maintaining local consistency.

\item We build PAV, a 460k-clip motion-centric AV dataset with automatic motion annotations, and show through extensive experiments that Tora3 consistently outperforms competitive baselines in video quality, trajectory faithfulness, and motion-sound coherence.
\end{itemize}

\section{Related work}

\subsection{Trajectory-Controlled Video Generation}
Trajectory-controlled video generation has attracted increasing attention for fine-grained manipulation of object motion and scene dynamics. Prior works~\cite{li2025image, shi2024motion, yin2023dragnuwa, wang2024motionctrl, li2025magicmotion, zhang2025tora, chu2025wan} inject motion cues such as optical flow~\cite{yin2023dragnuwa, shi2024motion}, point tracks~\cite{wang2024motionctrl, zhang2025tora, zhang2025tora2}, or temporally varying bounding boxes~\cite{dai2023animateanythingfinegrainedopendomain, li2025magicmotion} into diffusion models via dedicated encoders, adapters, or attention-based fusion. To represent complex motions, some methods~\cite{xiao20243dtrajmaster, wang2024levitor} explore 3D trajectories or depth-augmented keypoint maps, while others~\cite{geng2025motionpromptingcontrollingvideo, chu2025wan} use pretrained base models without architectural modifications for scalability. However, these mainly target video-only generation and do not address joint audio-video synthesis. In contrast, Tora3 extends trajectory guidance to unified audio-video generation, using trajectories as a shared kinematic prior for both motion control and audio-motion alignment.

\subsection{Audio-Video Generation}
Most prior unified audio-video generation methods~\cite{liu2025javisdit, ruan2023mm, wang2025av, xing2024seeing, zhang2025deepaudio} synthesize ambient sounds using dual-branch architectures, in which audio and video streams are modeled separately and fused through cross-attention or feature concatenation. Large-scale systems such as Veo3 show strong text-conditioned joint generation, and recent open-source models such as UniVerse-1~\cite{wang2025universe}, Ovi~\cite{low2025ovi}, and LTX-2~\cite{hacohen2026ltx2efficientjointaudiovisual} largely follow the same dual-backbone fusion paradigm. JoVA~\cite{huang2025jova} further enables direct cross-modal interaction for human-centric content. Nevertheless, these often struggle to keep audio well aligned with object motion and interaction timing. In contrast, Tora3 uses object trajectories as a shared kinematic prior, combining trajectory-guided motion control in the video branch with kinematic-audio alignment in the audio branch, improving motion-sound coherence by aligning sound events and intensity changes with trajectory-level motion dynamics.

\begin{figure*}[!t]
    \centering
    \includegraphics[width=1\textwidth]{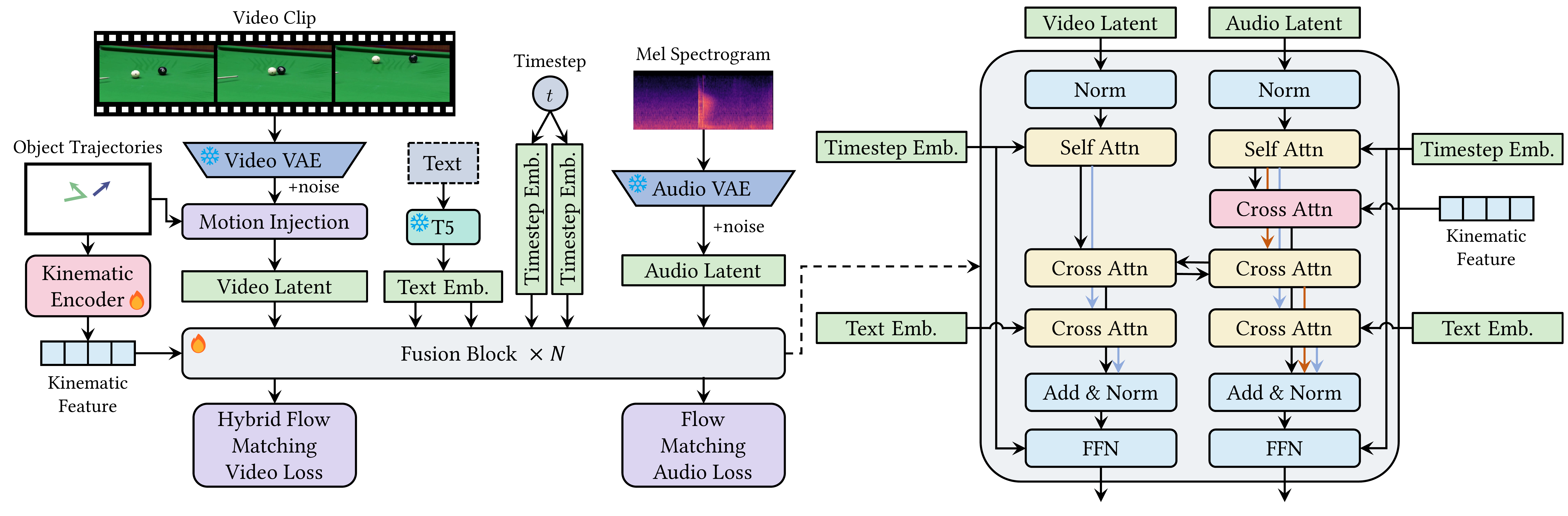}
    \caption{The overall architecture of Tora3. Object trajectories provide a shared kinematic prior for both video and audio generation. The video branch uses trajectory-aligned motion injection, while the audio branch incorporates trajectory-derived kinematic features to better align sound with motion. Hybrid flow matching further improves trajectory-guided generation by balancing motion fidelity and local coherence.}
    \Description{A block diagram of the Tora3 architecture showing a twin diffusion transformer with a video branch and an audio branch. Object trajectories feed a shared kinematic prior that supplies trajectory-aligned motion injection to the video branch and trajectory-derived kinematic features to the audio branch.}

    \label{f:pipeline}   
\end{figure*}

\section{Methods}

\subsection{Overview}

Tora3 is an audio-video generation framework built on a twin diffusion transformer architecture with separate video and audio DiT backbones, following Ovi~\cite{low2025ovi}. It uses object trajectories as a shared kinematic prior across both modalities and consists of three components: \emph{trajectory-aligned motion representation} for video, \emph{kinematic-audio alignment} module driven by second-order kinematic states for audio, and \emph{hybrid flow matching} for preserving motion fidelity and local coherence. The overall framework is illustrated in Figure~\ref{f:pipeline}.

\subsection{Trajectory-aligned Motion Representation}

Following Tora~\cite{zhang2025tora} and Tora2~\cite{zhang2025tora2}, we use object trajectories to represent both local motion cues and longer-range temporal evolution. A common strategy in controllable video generation is to first encode trajectories with an auxiliary motion encoder and then inject the resulting features into the generative backbone. However, adopting such a design in a unified audio-video framework is suboptimal. Since both audio and video branches are large DiT backbones, an additional motion encoder would further entangle optimization and make it harder to preserve a stable and transferable~\cite{qiang2025transferability} motion representation across modalities. Moreover, extra encoding layers may attenuate sparse trajectory signals before they reach the backbone, reducing motion fidelity.

To avoid these issues, Tora3 directly maps trajectories into the latent space of the video backbone without introducing a dedicated motion encoder.
We follow a standard image-to-video pipeline. Let
$z \in \mathbb{R}^{h \times w \times d}$ denote the latent of the first frame, obtained
by a pretrained VAE encoder~\cite{kingma2022autoencodingvariationalbayes}, where $h$
and $w$ are the latent spatial dimensions and $d$ is the channel dimension. Let
$x_0 \in \mathbb{R}^{f \times h \times w \times d}$ denote the clean video latent
sequence, where $f$ is the number of latent frames and the first frame corresponds to
the conditioning image. Let $\widetilde{\mathcal{T}}$
denote the full object trajectories in pixel space. We apply temporal average pooling
to $\widetilde{\mathcal{T}}$ to match the sampled video frames, obtaining
$\mathcal{T} \in \mathbb{R}^{f \times N \times 2}$. We further map $\mathcal{T}$ to
latent coordinates by scaling with the VAE downsampling factor and applying
nearest-neighbor rounding, yielding $\mathcal{T}'$, where each
$\mathcal{T}'_{i,n} = (p_{i,n}, q_{i,n}) \in \{0,\dots,h-1\}\times\{0,\dots,w-1\}$.
Here, $p_{i,n}$ and $q_{i,n}$ denote the row and column index, respectively, of
object $n$ at frame $i$ on the latent grid. We construct a trajectory-conditioned latent
$x^{\mathrm{traj}} \in \mathbb{R}^{f \times h \times w \times d}$ by propagating
first-frame object features along the trajectories. For each object
trajectory, we replace the latent feature at the trajectory-aligned location in frame
$i$ with the corresponding first-frame feature:
\begin{equation}
    x^{\mathrm{traj}}[i, p_{i,n}, q_{i,n}, :] = z[p_{0,n}, q_{0,n}, :],
    \quad i \in \{1,\dots,f-1\}.
\end{equation}
If multiple trajectories map to the same latent location, we randomly select one. All non-trajectory locations in \(x^{\mathrm{traj}}\) remain zero, except in the first frame, where they are overwritten by \(z\).
This design is motivated by two empirical observations: VAE latents are spatially smooth, so nearby locations share similar local appearance statistics~\cite{DBLP:conf/cvpr/RombachBLEO22}; and relocating latent features to nearby trajectory-aligned positions often preserves local object cues~\cite{DBLP:conf/eccv/WuLGZHZSLGZ24}. Together, these make first-frame latent propagation a direct and effective way to inject motion cues into the native latent space, providing an explicit trajectory-aligned conditioning signal while avoiding the additional feature transformation introduced by a separate motion encoder.

\subsection{Kinematic-Audio Alignment}

Semantic conditioning alone is insufficient for realistic motion-dependent audio synchronization. In monaural audio generation, motion-sound alignment depends heavily on temporal structure, spectral texture, and amplitude evolution, all of which are often correlated with object motion. Without explicit kinematic cues, the model must infer these relations implicitly from visual features, often leading to inaccurate timing and mismatched intensity.

\noindent \textbf{Kinematic Feature Representation.}
To introduce explicit motion-aware structure into audio generation, we condition the audio branch on compact kinematic features derived from object trajectories. These form a second-order kinematic state, since an object's motion can be described by its position, velocity, and acceleration, providing cues about when interactions occur, how strong they are, and how acoustic energy may evolve over time.

Given temporal-downsampled trajectories $\mathcal{T} \in \mathbb{R}^{f \times N \times 2}$, we normalize the 2D coordinates by the longer side of the image and denote the normalized position of object $n$ at frame $i$ as $\bm{r}_{i,n} \in [0, 1]^2$. These provide a coarse spatial anchor and scene context, such as whether an object is near the ground or suspended in the air, helping infer the interaction type and associated sound.

We estimate velocity and acceleration using second-order central differences for interior frames, with forward and backward differences at the boundaries. Let $\tau = 1/\text{fps}$ denote the time interval between consecutive frames. The velocity and acceleration at frame $i$ for trajectory point $n$ are approximated as
\begin{equation}
    \bm{v}_{i,n}  \approx  \frac{\bm{r}_{i+1,n} - \bm{r}_{i-1,n}}{2\tau},
\end{equation}
\begin{equation}
    \bm{a}_{i,n} \approx  \frac{\bm{r}_{i+1,n} - 2\bm{r}_{i,n} + \bm{r}_{i-1,n}}{\tau^2},
\end{equation}
valid for interior frames $i \in \{1, \dots, f-2\}$. This provides stable, second-order accurate estimates of trajectory-level motion. We explicitly retain both vector components and magnitudes to disentangle direction from strength.

The velocity vector characterizes the mode of interaction, for example, predominantly vertical motion for falling and impact, or horizontal motion for sliding and rolling, which can be related to different sound patterns. Its magnitude $\|\bm{v}_{i,n}\|_2$ reflects motion intensity. Acceleration $\bm{a}_{i,n}$ is particularly informative for contact-dominant events such as impacts, abrupt stops, or braking, where rapid changes in motion often correlate with stronger and more salient acoustic responses. Its direction indicates whether motion is increasing or decelerating, while its magnitude $\|\bm{a}_{i,n}\|_2$ provides a useful cue for event strength.

We use $\|\bm{v}_{i,n}\|_2$ and $\|\bm{a}_{i,n}\|_2$ as explicit motion-intensity cues that help align audio amplitude and temporal structure with the underlying motion, improving optimization and helping the model distinguish sound behaviors such as impact-like transients versus sustained scraping or rolling.

We combine these terms into an 8D kinematic feature for each object and frame:
\begin{equation}
    \bm{\phi}_{i,n} = \left[ \bm{r}_{i,n}^{\top}, \bm{v}_{i,n}^{\top}, \bm{a}_{i,n}^{\top}, \|\bm{v}_{i,n}\|_2, \|\bm{a}_{i,n}\|_2 \right]^{\top} \in \mathbb{R}^8.
\end{equation}
This representation explicitly exposes the key kinematic factors relevant to motion-aware sound generation, namely spatial context, interaction mode, and motion intensity. Since these terms may span several orders of magnitude, we apply signed log-compression before encoding. Let
\begin{equation}
    \bm{d}_{i,n} = \left[ \bm{v}_{i,n}^{\top}, \bm{a}_{i,n}^{\top}, \|\bm{v}_{i,n}\|_2, \|\bm{a}_{i,n}\|_2 \right]^{\top},
\end{equation}
\begin{equation}
    \tilde{\bm{d}}_{i,n} = \mathrm{sign}(\bm{d}_{i,n}) \odot \log_{10}(1 + |\bm{d}_{i,n}|).
\end{equation}
We then encode the normalized kinematic feature as
\begin{equation}
    \tilde{\bm{\phi}}_{i,n} = \mathcal{E}_k\!\left(\mathrm{Norm}\!\left(\left[\bm{r}_{i,n}^{\top}, \tilde{\bm{d}}_{i,n}^{\top}\right]^{\top}\right)\right),
\end{equation}
yielding latent kinematic tokens in $\mathbb{R}^d$ that match the embedding dimension of the audio transformer, where $\mathcal{E}_k$ denotes a three-layer MLP.

\noindent \textbf{Audio-Kinematic Fusion.}
The encoded kinematic tokens $\bm{H}_{\text{kin}} = \{\tilde{\bm{\phi}}_{i,n}\} \in \mathbb{R}^{f \times N \times d}$ are injected into the audio branch of the twin-DiT to explicitly modulate audio generation with motion, analogous to injecting compact guidance vectors into large pretrained models~\cite{zhang2026enhancing}. Within each audio transformer block, an auxiliary cross-attention layer is added after the self-attention layer, with audio latent states as queries and kinematic tokens as keys and values.
We project $\bm{H}_{\text{kin}}$ to key and value tensors. For clarity, we omit the multi-head notation:
\begin{equation}
\bm{K}_{\text{kin}} = \mathrm{Norm}(\bm{H}_{\text{kin}} \bm{W}_{\text{kin}}^{(K)}), \quad
\bm{V}_{\text{kin}} = \bm{H}_{\text{kin}} \bm{W}_{\text{kin}}^{(V)}.
\end{equation}
To preserve temporal alignment, we apply RoPE~\cite{DBLP:journals/ijon/SuALPBL24} to the temporal dimension of $\bm{K}_{\text{kin}}$, reshape $\bm{K}_{\text{kin}}$ and $\bm{V}_{\text{kin}}$ to $\mathbb{R}^{(fN) \times d}$, and apply RoPE to audio latents $\bm{H}_a \in \mathbb{R}^{L_a \times d}$, with temporal indices scaled by $f/L_a$, following Ovi.
The kinematic-aware update $\Delta \bm{H}_a$ is computed as:
\begin{align}
    \bm{Q}_a &= \mathrm{Norm}(\bm{H}_a \bm{W}_a^{(Q)}), \\
    \Delta \bm{H}_a &= \mathrm{Softmax}\!\left(\frac{\bm{Q}_a \bm{K}_{\text{kin}}^{\top}}{\sqrt{d}}\right) \bm{V}_{\text{kin}}.
\end{align}
To prevent kinematic cues from dominating semantic text conditions during early training and to preserve the pretrained model's semantic capability while adapting to the new motion conditioning~\cite{qiang2026plasticity}, we further employ a learnable gating mechanism. The final hidden state $\bm{H}_a'$ is updated through a residual connection modulated by a sigmoid-gated scalar $\sigma(\gamma)$, where $\gamma$ is a learnable parameter initialized to a negative value:
\begin{equation}
    \bm{H}_a' = \bm{H}_a + \sigma(\gamma) \cdot \Delta \bm{H}_a.
\end{equation}
This allows Tora3 balance semantic context and motion-aware conditioning, producing audio that is both semantically appropriate and better aligned with the visual motion.

\begin{table*}[!t]
\setlength{\tabcolsep}{3.5pt}
\centering
\caption{Comparison on video quality, audio quality, text alignment, motion control, and audio-video synchronization, including event-level and intensity-level motion-sound coherence metrics.}
\begin{tabular}{cccccccccccccc}
\toprule
 & &\multicolumn{2}{c}{Video Quality} & \multicolumn{4}{c}{Audio Quality} &  \multicolumn{2}{c}{Text Alignment} & \multicolumn{3}{c}{AV Synchronization} & Motion Control \\
\cmidrule(lr){3-4} \cmidrule(lr){5-8} \cmidrule(lr){9-10} \cmidrule(lr){11-13} \cmidrule(lr){14-14}

Method & \# Params  & AS $\uparrow$ & FVD $\downarrow$ & CE $\uparrow$ & CU $\uparrow$ & PC $\downarrow$ & PQ $\uparrow$ & CLAP $\uparrow$ & CLIP-T $\uparrow$ & FGAS $\uparrow$ & ETE $\downarrow$ & MAIC $\uparrow$ & TrajError $\downarrow$\\ 
\midrule
LTX-2~\cite{hacohen2026ltx2efficientjointaudiovisual} & 22.16B & 4.31 & 989.6 & 3.28 & 6.17 & 2.43 & 6.73 & 0.31 & 0.29 & 0.187 & 0.284 & 0.41 & -\\ 
Ovi~\cite{low2025ovi} & 11.66B & 4.40 & 887.7 & \underline{3.30} & 6.01 & 1.85 & 6.44 & 0.43 & 0.30 & 0.156 & 0.301 & 0.37 & -\\ 
MOVA~\cite{openmossteam2026movascalablesynchronizedvideoaudio} & 30.00B& \textbf{4.63} & 849.8 & 3.05 & \underline{6.31} & \textbf{1.77} & \underline{6.95} & \textbf{0.46} & \textbf{0.31} & 0.201 & 0.236 & 0.49 & -\\ 
AVControl~\cite{benyosef2026avcontrolefficientframeworktraining} & 22.32B & 4.52 & \underline{829.6} & 3.29 & 6.22 & 2.18 & 6.79 & 0.39 & 0.30 & \underline{0.209} & \underline{0.214} & \underline{0.55} & \underline{19.95}\\
\textbf{Tora3} & 12.25B & \underline{4.61} & \textbf{784.1} & \textbf{3.34} & \textbf{6.43} & \underline{1.81} & \textbf{7.09} & \underline{0.44} & \textbf{0.31} & \textbf{0.234} & \textbf{0.181} & \textbf{0.63} & \textbf{12.13}\\

\bottomrule
\end{tabular}
\label{t:compare}
\end{table*}

\subsection{Hybrid Flow Matching}

Directly reusing first-frame latents along trajectories is effective for injecting motion cues, but applying the same construction uniformly to all spatial locations can be limiting. We observe that different regions play different roles during denoising. Trajectory-conditioned regions benefit from stronger appearance anchoring to preserve motion guidance, whereas non-trajectory regions benefit from the standard flow to maintain flexible scene refinement. Motivated by this observation, we propose \textit{Hybrid Flow Matching}, which modifies the probability flow inside the trajectory region $\Omega_{\text{traj}} =\left\{(i, p_{i,n}, q_{i,n})\;\middle|\;i \in \{0,\dots,f-1\},\ n \in \{0,\dots,N-1\}\right\}.$ while preserving the standard flow outside it.

Outside $\Omega_{\text{traj}}$, we follow the standard Flow Matching construction on the
clean video latent $x_0$:
\begin{equation}
    x_t = (1-t)x_0 + t \epsilon, \qquad
    v = \epsilon - x_0,
\end{equation}
where $t \in [0,1]$ denotes the flow matching timestep, $\epsilon \sim \mathcal{N}(0, I)$
denotes the standard Gaussian endpoint, and $v$ denotes the target
velocity field for unconstrained regions.

Inside $\Omega_{\text{traj}}$, we replace the standard noise endpoint with the
trajectory-conditioned latent $x^{\mathrm{traj}}$, yielding
\begin{equation}
    x_t = (1-t)x_0 + t x^{\mathrm{traj}}, \qquad
    v = x^{\mathrm{traj}} - x_0.
\end{equation}
Intuitively, this preserves the standard denoising behavior in unconstrained regions
while anchoring trajectory-conditioned regions to the injected motion prior.

To implement this hybrid construction, we use the binary indicator mask $M \in \{0,1\}^{f \times h \times w}$ associated with $\Omega_{\text{traj}}$, defined by $M[i,p,q]=\mathbf{1}[(i,p,q)\in\Omega_{\text{traj}}]$.
The resulting latent input and target velocity field are defined as
\begin{equation}
    x_t = (1-M) \odot \big((1-t)x_0 + t \epsilon\big) + M \odot \big((1-t)x_0 + t x^{\mathrm{traj}}\big),
\end{equation}

\begin{equation}
    v = (1-M) \odot (\epsilon - x_0) + M \odot (x^{\mathrm{traj}} - x_0).
\end{equation}
Here, the mask $M$ is broadcast along the channel dimension to $\mathbb{R}^{f \times h \times w \times d}$. At inference time, we use the same trajectory-conditioned latent construction to initialize the flow-matching process at \(t=1\), with \(x_1 = (1-M)\odot \epsilon + M \odot x^{\mathrm{traj}}\).

This ensures that the latent input at each spatial location is consistent with the corresponding velocity target used for supervision, helping preserve trajectory-guided motion while reducing local inconsistencies around trajectory-conditioned regions.

During training, we keep the audio objective unchanged from Ovi and only replace its original video loss with a region-balanced objective to better balance the sparse trajectory region and its complement. To ensure smooth boundary handling, we use a soft mask $M_{\text{soft}}$ obtained by Gaussian blurring the binary mask $M$. The modified video loss is defined as
\begin{equation}
    \mathcal{L}_{\text{video}} = \lambda_{\text{out}} \mathcal{L}_{\text{out}} + \lambda_{\text{traj}} \mathcal{L}_{\text{traj}},
\end{equation}
where $\lambda_{\text{out}} = \lambda_{\text{traj}} = 0.5$. The loss terms are defined as
\begin{align}
    \mathcal{L}_{\text{out}} &= \frac{\sum \big((1-M_{\text{soft}}) \odot (\hat{v} - v)^2\big)}{\sum (1-M_{\text{soft}}) + \delta}, \\
    \mathcal{L}_{\text{traj}} &= \frac{\sum \big(M_{\text{soft}} \odot (\hat{v} - v)^2\big)}{\sum M_{\text{soft}} + \delta},
\end{align}
where $\delta = 10^{-8}$ is a small constant for numerical stability.
This equal weighting prevents the sparse trajectory region from being overwhelmed by the dominant non-trajectory area, allowing the model learn localized motion control and global appearance more evenly. Following Ovi, the final training objective combines the video and audio losses with weights $0.85$ and $0.15$, respectively:
\begin{equation}
    \mathcal{L}_{\text{final}} = 0.85\,\mathcal{L}_{\text{video}} + 0.15\,\mathcal{L}_{\text{audio}}^{\text{Ovi}},
\end{equation}
where $\mathcal{L}_{\text{audio}}^{\text{Ovi}}$ denotes the original audio loss used in Ovi.

\section{Experiments}
\subsection{Experimental Setup}

\noindent \textbf{Datasets.}
We construct PAV from filtered high-quality subsets of VGGSound~\cite{chen2020vggsoundlargescaleaudiovisualdataset}, ACAV-100M~\cite{lee2021acav100mautomaticcurationlargescale}, OpenVid1M~\cite{nan2025openvid1mlargescalehighqualitydataset}, Pexels~\cite{pexel}, and in-house collected data. To improve motion relevance, we use Qwen3-VL~\cite{bai2025qwen3vltechnicalreport} to automatically filter clips whose main object exhibits basic motion patterns such as translation, rotation, sliding, parabolic motion, oscillation, scale change, or deformation. For the retained clips, we apply SAM2~\cite{ravi2024sam2segmentimages} to segment all objects in the first frame, then run CoTracker3~\cite{karaev2024cotracker3simplerbetterpoint} initialized from each object's centroid to obtain trajectory annotations, resulting in 460k clips providing scalable kinematic supervision rather than full physical measurements. We further use Qwen3-VL-8B-Instruct and Qwen3-Omni-Captioner~\cite{xu2025qwen3omnitechnicalreport} to generate paired video and audio descriptions. For quantitative evaluation, we use 50 representative videos covering diverse motion patterns and scene types.

\noindent \textbf{Metrics.}
We evaluate generation quality from four aspects: video quality, audio quality, text alignment, and audio-video synchronization. For video, we report Fréchet Video Distance (FVD)~\cite{unterthiner2019accurategenerativemodelsvideo}, Aesthetic Score (AS), computed as the average of MANIQA~\cite{yang2022maniqamultidimensionattentionnetwork} and MUSIQ~\cite{ke2021musiqmultiscaleimagequality}, and Trajectory Error (TrajError), defined as the L2 distance between the conditioning trajectories and the object tracks estimated from generated videos, following~\cite{zhang2025tora}. For audio, we adopt AudioBox-Aesthetics~\cite{tjandra2025metaaudioboxaestheticsunified}, including Content Enjoyment (CE), Content Usefulness (CU), Production Complexity (PC), and Production Quality (PQ). For text alignment, we compute CLIP~\cite{radford2021learningtransferablevisualmodels} and CLAP~\cite{wu2024largescalecontrastivelanguageaudiopretraining} scores for text-to-video and text-to-audio consistency. For audio-video synchronization, we use FGAS in PhyAVBench~\cite{xie2025phyavbenchchallengingaudiophysicssensitivity}, the cosine similarity between video frames and synchronized audio features in the latent space of CAV-MAE Sync~\cite{gong2023contrastiveaudiovisualmaskedautoencoder}. To further assess physical coherence at the motion-sound level, we add two metrics: Event Timing Error (ETE) measures the time difference between trajectory-derived events and corresponding audio onsets, and Motion-Audio Intensity Correlation (MAIC) computes the Pearson correlation between a trajectory-derived motion-intensity signal and the temporally aligned audio energy envelope. We evaluate ETE on clips with discrete events and MAIC on clips with sustained motion and continuous sound evolution.

\noindent \textbf{Implementation Details.} The model is initialized from the pretrained Ovi checkpoint and trained for 30k steps on 32 NVIDIA A100 GPUs with a global batch size of 32. We use AdamW ($\beta_1=0.9$, $\beta_2=0.999$, weight decay 0.01, learning rate $4\times10^{-5}$), with BF16 mixed precision and gradient clipping at 1.0 for stability. Kinematic features are normalized using dataset-level statistics from 5,000 randomly sampled training examples, and we apply trajectory condition dropout with probability $p=0.05$ to improve robustness. For $M_{\text{soft}}$, the binary trajectory mask is smoothed with a Gaussian kernel of $\sigma=0.5$ to encourage smooth transitions between trajectory-conditioned regions and their surroundings. The sigmoid-gated scalar $\gamma$ is initialized to $-10$.

\subsection{Main Results}

\begin{figure*}[!t]
    \centering
    \includegraphics[width=1\textwidth]{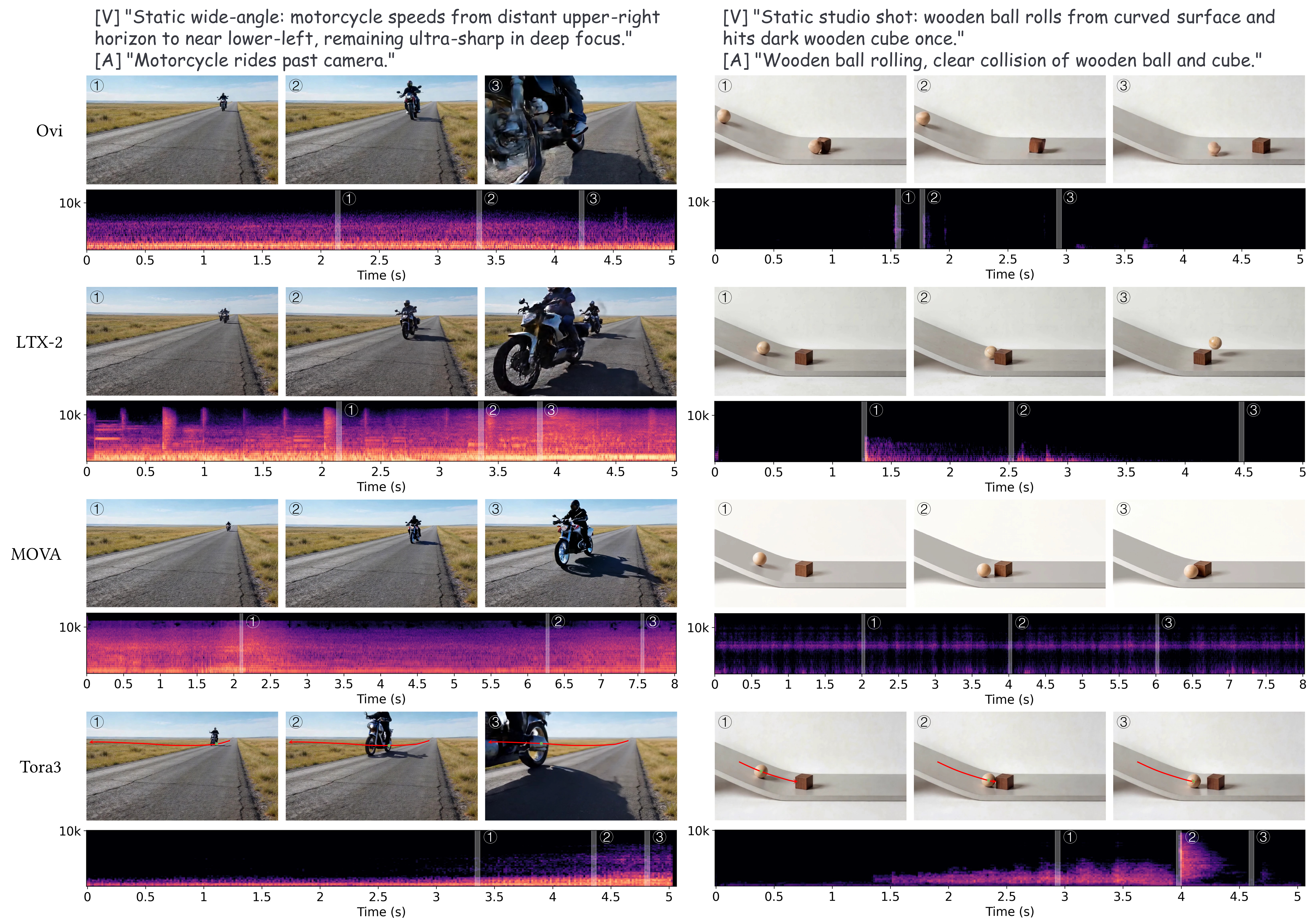}
    \caption{Qualitative comparison. Tora3 produces more stable visual motion and more coherent audio in both examples, including motion-consistent loudness variation for the moving motorcycle and better synchronized rolling and collision sounds for the rolling ball.}
    \Description{Side-by-side qualitative comparison between Tora3 and baseline methods on a moving motorcycle example and a rolling ball example, with video frames and audio waveforms showing that Tora3 yields more stable motion and better synchronized loudness, rolling, and collision sounds.}
    \label{f:compare}
\end{figure*}

Table~\ref{t:compare} compares Tora3 with four competitive baselines, LTX-2~\cite{hacohen2026ltx2efficientjointaudiovisual}, Ovi~\cite{low2025ovi}, MOVA~\cite{openmossteam2026movascalablesynchronizedvideoaudio}, and AVControl~\cite{benyosef2026avcontrolefficientframeworktraining}, across video quality, audio quality, text alignment, audio-video synchronization, and motion control. AVControl is a concurrent work on motion-conditioned audio-video generation, included here for direct comparison. Overall, Tora3 achieves the best trade-off of generation quality and motion-sound coherence: it delivers the best video realism with the lowest FVD, strong overall audio quality, and the highest FGAS, indicating improved motion-sound synchronization. Tora3 also achieves the lowest ETE and the highest MAIC, showing that it more accurately aligns salient audio events with motion and contact cues and better matches audio intensity to motion strength. Meanwhile, it remains highly competitive in text alignment, matching the best CLIP-T score with a strong CLAP score.

Compared with AVControl, Tora3 is particularly strong in motion-sound coherence. AVControl mainly uses trajectories as a control signal for the joint backbone, whereas Tora3 uses them as a shared kinematic prior for both modalities. Trajectory-aligned motion injection improves video motion fidelity, and kinematic-audio alignment explicitly links motion states to sound, yielding better FGAS, ETE, and MAIC. These gains support our central design: trajectories couple visual motion and acoustic generation under a shared trajectory-grounded structure.

Figure~\ref{f:compare} shows the qualitative advantages of Tora3 over competitive baselines. In the motorcycle example, baselines either degrade visual quality or fail to produce motion-consistent loudness variation, whereas Tora3 maintains stable visuals and more coherent audio dynamics. In the rolling-ball example, baselines struggle with rolling, collision, and post-impact motion together with synchronized sound, often producing unrealistic motion, incorrect impact timing, or audio artifacts. Tora3 models these events more faithfully with better synchronized rolling and impact sounds.

\begin{figure}[!t]
    \centering
    \includegraphics[width=0.45\textwidth]{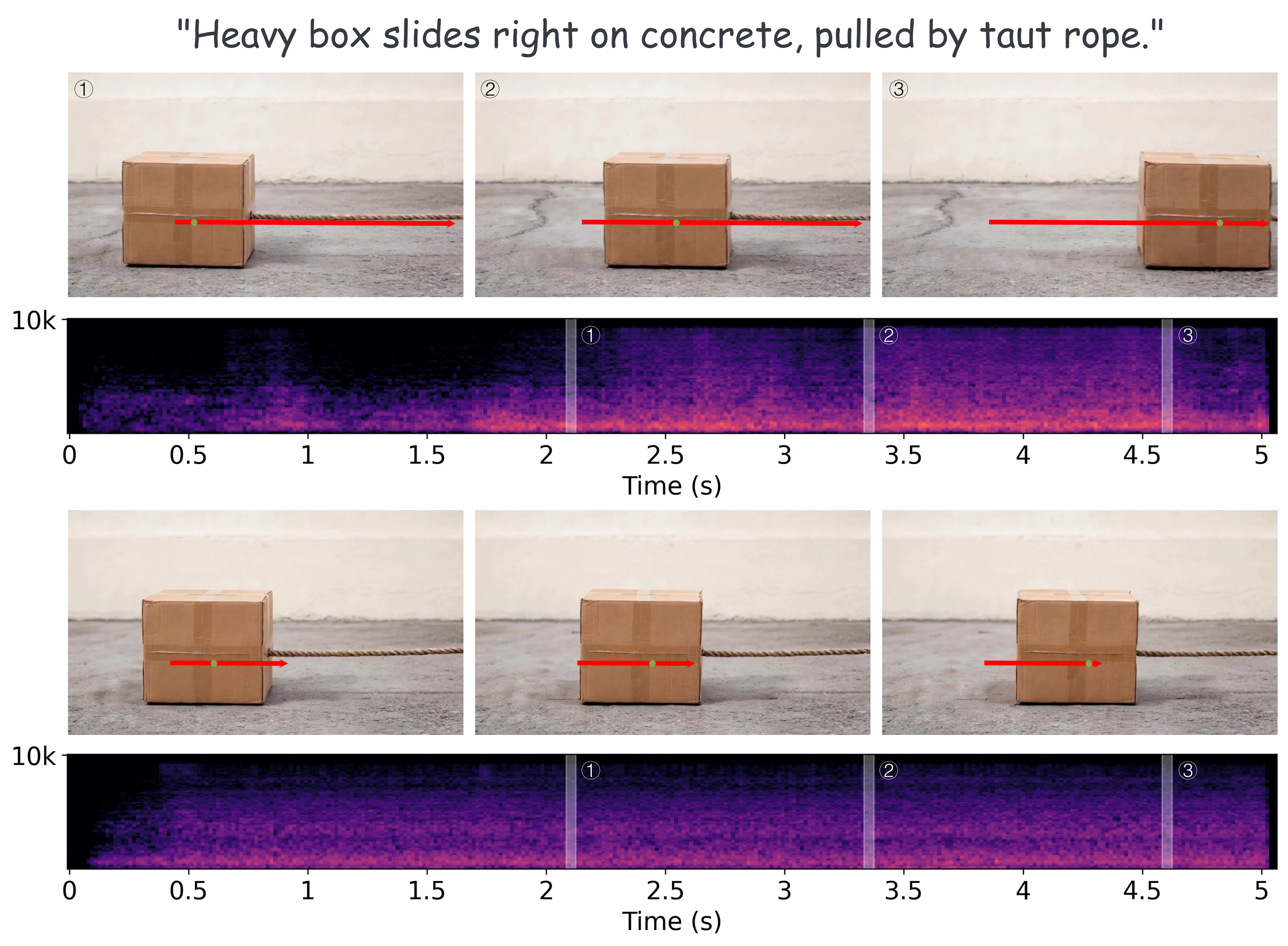}
    \caption{Effect of motion speed on generated audio. Tora3 generates audio that varies consistently with motion speed.}
    \Description{An illustration of a box being pulled at different speeds together with the corresponding generated audio waveforms, showing that faster motion produces louder sound with subtle timbral differences compared with slower motion.}
    \label{f:speed}
\end{figure}

Figure~\ref{f:speed} shows that Tora3 produces speed-dependent audio variations. When the box is pulled faster, the generated sound is louder and exhibits subtle timbral differences compared with slower motion. This highlights the advantage of trajectories as a shared kinematic prior, allowing motion changes to be reflected more consistently in both visual dynamics and generated sound.

\subsection{Ablation Studies}
\label{ab:sec}

\begin{sloppypar}
\noindent \textbf{Shared kinematic prior across modalities.}
Table~\ref{t:shared_backbone} studies whether trajectory-derived kinematic conditioning should be applied to the video branch, the audio branch, or both. Applying it only to the video branch mainly improves visual generation, yielding the best Aesthetic Score and clearly better video realism, whereas applying it only to the audio branch more directly improves audio quality and motion-sound coherence, as reflected by higher PQ, FGAS, and MAIC and lower ETE. These complementary effects show that trajectory injection improves motion faithfulness in the video branch, while kinematic conditioning improves event timing and intensity alignment in the audio branch. Used together, they produce the strongest overall results, confirming that trajectories are most effective as a shared kinematic prior across modalities rather than isolated controls for either branch alone.
\end{sloppypar}

\begin{table}[!t]
\centering
\caption{Effect of using trajectories as a shared kinematic prior for both modalities.}
\setlength{\tabcolsep}{4pt}
\small
\begin{tabular}{lcccccc}
\toprule
Setting & AS $\uparrow$ & FVD $\downarrow$ & PQ $\uparrow$ & FGAS $\uparrow$ & ETE $\downarrow$ & MAIC $\uparrow$ \\
\midrule
Neither     & 4.39 & 854.7 & 6.75 & 0.173 & 0.296 & 0.39 \\
Video only  & \textbf{4.51} & 823.6 & 6.81 & 0.198 & 0.247 & 0.46 \\
Audio only  & 4.42 & 845.2 & 6.89 & 0.209 & 0.221 & 0.61 \\
Both (full) & 4.47 & \textbf{811.8} & \textbf{6.93} & \textbf{0.225} & \textbf{0.193} & \textbf{0.66} \\
\bottomrule
\end{tabular}
\label{t:shared_backbone}
\end{table}

\noindent \textbf{Motion representation designs for video-branch injection.}
Table~\ref{t:ablation1} compares three motion injection strategies for the video branch: a Tora-inspired design using an auxiliary motion encoder and fuser, a WanMove-inspired design that concatenates latent noise with trajectory-guided first-frame propagated features along the channel dimension, and our Tora3 motion representation. Tora3 consistently achieves the best overall performance, yielding the highest visual quality, the strongest audio-video synchronization, the lowest event timing error, and the lowest trajectory error, while requiring no additional parameters. This advantage likely stems from the fact that: unlike prior approaches that transform sparse control signals into separate feature spaces, Tora3 injects motion directly in the native latent space by reusing first-frame features along the prescribed trajectories. This leads to cleaner and more reliable motion control, improving trajectory following in the video branch and providing more stable motion cues to the audio branch through cross-modal interaction.

\noindent \textbf{Kinematic state components.}
Table~\ref{t:ablation2} studies the effect of progressively enriching the kinematic conditioning used by the audio branch. Starting from no conditioning, adding position and velocity already improves both synchronization and motion-sound coherence. Incorporating acceleration further strengthens performance, especially in FGAS, ETE, and MAIC. The full second-order kinematic state, which additionally includes velocity and acceleration magnitudes, achieves the best performance across all metrics. This trend matches the role of each component: position provides coarse scene context, velocity indicates how objects move over time, and acceleration is particularly informative for abrupt changes associated with impact-like events, while the magnitudes expose motion strength for better calibrating audio intensity and temporal salience. These monotonic improvements support our formulation of trajectory-derived kinematic states as a compact and effective interface between trajectories and audio generation.

\begin{table}[!t]
\setlength{\tabcolsep}{2pt}
\centering
\caption{Ablation on motion representation designs for video-branch injection. Tora-style and WanMove-style denote variants inspired by Tora~\cite{zhang2025tora} and WanMove~\cite{chu2025wan}, respectively.}
\small
\begin{tabular}{lccccc}
\toprule
Method & AS $\uparrow$ & FGAS $\uparrow$ & ETE $\downarrow$ & TrajError $\downarrow$ & \# Params \\ 
\midrule
Tora-style~\cite{zhang2025tora} & 4.49 & 0.179 & 0.271 & 17.06 & 12.65B (+0.99B) \\ 
WanMove-style~\cite{chu2025wan} & 4.44 & 0.184 & 0.251 & 13.91 & 11.66B (+590K) \\
\textbf{Ours} & \textbf{4.51} & \textbf{0.198} & \textbf{0.247} & \textbf{13.03} & \textbf{11.66B (+0)} \\
\bottomrule
\end{tabular}
\label{t:ablation1}
\end{table}

\begin{table}[!t]
\setlength{\tabcolsep}{4pt}
\centering
\caption{Effect of different kinematic state components on audio generation and motion-sound coherence.}
\begin{tabular}{lcccc}
\toprule
Kinematic Signal & PQ $\uparrow$ & FGAS $\uparrow$ & ETE $\downarrow$ & MAIC $\uparrow$ \\ 
\midrule
None & 6.75 & 0.173 & 0.296 & 0.39 \\ 
$\bm{r}+\bm{v}$ & 6.79 & 0.191 & 0.254 & 0.48 \\ 
$\bm{r}+\bm{v}+\bm{a}$ & 6.86 & 0.202 & 0.229 & 0.56 \\
$\bm{r}+\bm{v}+\bm{a}+\|\bm{v}\|_2+\|\bm{a}\|_2$
 & \textbf{6.89} & \textbf{0.209} & \textbf{0.221} & \textbf{0.61}\\
\bottomrule
\end{tabular}
\label{t:ablation2}
\end{table}

\begin{table}[!t]
\setlength{\tabcolsep}{3pt}
\centering
\caption{Effect of hybrid flow matching on video quality, trajectory following, and overall AV synchronization.}
\begin{tabular}{ccccc}
\toprule
Method & AS $\uparrow$ & FVD $\downarrow$ & TrajError $\downarrow$ & FGAS $\uparrow$ \\ 
\midrule
w/o Hybrid flow matching & 4.47 & 811.8 & 12.94 & 0.225\\
w/ Hybrid flow matching & \textbf{4.61} & \textbf{784.1} & \textbf{12.13} & \textbf{0.234} \\ 
\bottomrule
\end{tabular}
\label{t:ablation3}
\end{table}

\noindent \textbf{Hybrid flow matching.}
Table~\ref{t:ablation3} shows that hybrid flow matching consistently improves perceptual and distributional video quality, trajectory faithfulness, and overall audio-video synchronization, raising the Aesthetic Score from 4.47 to 4.61, reducing FVD from 811.8 to 784.1, and lowering TrajError from 12.94 to 12.13. These results support our region-aware construction: stronger appearance anchoring in trajectory-conditioned regions improves motion fidelity, while preserving the standard flow elsewhere maintains local visual coherence.

\section{Conclusion}

We presented Tora3, a trajectory-guided framework for audio-video generation that improves physical coherence by using object trajectories as a shared kinematic prior for video and audio synthesis. It integrates a trajectory-aligned motion representation, a second-order kinematics-based audio alignment module, and a hybrid flow matching objective for robust motion modeling and local consistency. We also introduced PAV, a large-scale audio-video dataset with motion-centric annotations. Extensive experiments show that Tora3 improves visual realism, audio quality, motion-sound coherence, and audio-video synchronization over strong open-source baselines, while remaining competitive in text alignment. Future work includes extending trajectory-level coherence to richer physical factors such as material properties, contact dynamics, and 3D acoustic propagation.

\bibliographystyle{ACM-Reference-Format}
\balance
\bibliography{sample-base}

@String{Computer = "{IEEE} Computer" }

@String{Springer = "Springer-Verlag" }

@inproceedings{li2025image,
  title={{Image Conductor}: Precision control for interactive video synthesis},
  author={Li, Yaowei and Wang, Xintao and Zhang, Zhaoyang and Wang, Zhouxia and Yuan, Ziyang and Xie, Liangbin and Shan, Ying and Zou, Yuexian},
  booktitle={Proceedings of the AAAI Conference on Artificial Intelligence},
  volume={39},
  number={5},
  pages={5031--5038},
  year={2025}
}

@inproceedings{shi2024motion,
  title={{Motion-I2V}: Consistent and controllable image-to-video generation with explicit motion modeling},
  author={Shi, Xiaoyu and Huang, Zhaoyang and Wang, Fu-Yun and Bian, Weikang and Li, Dasong and Zhang, Yi and Zhang, Manyuan and Cheung, Ka Chun and See, Simon and Qin, Hongwei and others},
  booktitle={ACM SIGGRAPH 2024 Conference Papers},
  pages={1--11},
  year={2024}
}

@article{yin2023dragnuwa,
  title={{DragNUWA}: Fine-grained control in video generation by integrating text, image, and trajectory},
  author={Yin, Shengming and Wu, Chenfei and Liang, Jian and Shi, Jie and Li, Houqiang and Ming, Gong and Duan, Nan},
  journal={arXiv preprint arXiv:\href{https://arxiv.org/abs/2308.08089}{2308.08089}},
  year={2023}
}

@inproceedings{zhang2025tora,
  title={{Tora}: Trajectory-oriented diffusion transformer for video generation},
  author={Zhang, Zhenghao and Liao, Junchao and Li, Menghao and Dai, Zuozhuo and Qiu, Bingxue and Zhu, Siyu and Qin, Long and Wang, Weizhi},
  booktitle={Proceedings of the Computer Vision and Pattern Recognition Conference},
  pages={2063--2073},
  year={2025}
}

@inproceedings{wang2024motionctrl,
  title={{MotionCtrl}: A unified and flexible motion controller for video generation},
  author={Wang, Zhouxia and Yuan, Ziyang and Wang, Xintao and Li, Yaowei and Chen, Tianshui and Xia, Menghan and Luo, Ping and Shan, Ying},
  booktitle={ACM SIGGRAPH 2024 Conference Papers},
  pages={1--11},
  year={2024}
}

@inproceedings{xiao20243dtrajmaster,
  title={{3DTrajMaster}: Mastering {3D} trajectory for multi-entity motion in video generation},
  author={Xiao, FU and Liu, Xian and Wang, Xintao and Peng, Sida and Xia, Menghan and Shi, Xiaoyu and Yuan, Ziyang and Wan, Pengfei and Zhang, Di and Lin, Dahua},
  booktitle={The Thirteenth International Conference on Learning Representations},
  year={2025}
}

@article{chu2025wan,
  title={{Wan-Move}: Motion-controllable video generation via latent trajectory guidance},
  author={Chu, Ruihang and He, Yefei and Chen, Zhekai and Zhang, Shiwei and Xu, Xiaogang and WANG, Dingdong and Yi, Hongwei and Liu, Xihui and Zhao, Hengshuang and Liu, Yu and others},
  journal={Advances in Neural Information Processing Systems},
  volume={38},
  pages={404--432},
  year={2026}
}

@inproceedings{li2025magicmotion, title={{MagicMotion}: Controllable Video Generation with Dense-to-Sparse Trajectory Guidance}, url={http://dx.doi.org/10.1109/iccv51701.2025.01126}, DOI={10.1109/iccv51701.2025.01126}, booktitle={2025 IEEE/CVF International Conference on Computer Vision (ICCV)}, publisher={IEEE}, author={Li, Quanhao and Xing, Zhen and Wang, Rui and Zhang, Hui and Dai, Qi and Wu, Zuxuan}, year={2025}, month=Oct, pages={12112–12123} }

@article{liu2025javisdit,
  title={{JavisDiT}: Joint audio-video diffusion transformer with hierarchical spatio-temporal prior synchronization},
  author={Liu, Kai and Li, Wei and Chen, Lai and Wu, Shengqiong and Zheng, Yanhao and Ji, Jiayi and Zhou, Fan and Jiang, Rongxin and Luo, Jiebo and Fei, Hao and others},
  journal={arXiv preprint arXiv:\href{https://arxiv.org/abs/2503.23377}{2503.23377}},
  year={2025}
}

@inproceedings{ruan2023mm,
  title={{MM-Diffusion}: Learning multi-modal diffusion models for joint audio and video generation},
  author={Ruan, Ludan and Ma, Yiyang and Yang, Huan and He, Huiguo and Liu, Bei and Fu, Jianlong and Yuan, Nicholas Jing and Jin, Qin and Guo, Baining},
  booktitle={Proceedings of the IEEE/CVF Conference on Computer Vision and Pattern Recognition},
  pages={10219--10228},
  year={2023}
}

@inproceedings{wang2025av,
  title={{AV-DiT}: Taming image diffusion transformers for efficient joint audio and video generation},
  author={Wang, Kai and Deng, Shijian and Shi, Jing and Hatzinakos, Dimitrios and Tian, Yapeng},
  booktitle={Proceedings of the 33rd ACM International Conference on Multimedia},
  pages={10486--10495},
  year={2025}
}

@inproceedings{xing2024seeing,
  title={Seeing and hearing: Open-domain visual-audio generation with diffusion latent aligners},
  author={Xing, Yazhou and He, Yingqing and Tian, Zeyue and Wang, Xintao and Chen, Qifeng},
  booktitle={Proceedings of the IEEE/CVF Conference on Computer Vision and Pattern Recognition},
  pages={7151--7161},
  year={2024}
}

@article{zhang2025deepaudio,
  title={{DeepAudio-V1}: Towards Multi-Modal Multi-Stage End-to-End Video to Speech and Audio Generation},
  author={Zhang, Haomin and Liu, Chang and Zheng, Junjie and Chen, Zihao and Ding, Chaofan and Di, Xinhan},
  journal={arXiv preprint arXiv:\href{https://arxiv.org/abs/2503.22265}{2503.22265}},
  year={2025}
}

@article{wang2025universe,
  title={{UniVerse-1}: Unified Audio-Video Generation via Stitching of Experts},
  author={Wang, Duomin and Zuo, Wei and Li, Aojie and Chen, Ling-Hao and Liao, Xinyao and Zhou, Deyu and Yin, Zixin and Dai, Xili and Jiang, Daxin and Yu, Gang},
  journal={arXiv preprint arXiv:\href{https://arxiv.org/abs/2509.06155}{2509.06155}},
  year={2025}
}

@article{low2025ovi,
  title={{Ovi}: Twin backbone cross-modal fusion for audio-video generation},
  author={Low, Chetwin and Wang, Weimin and Katyal, Calder},
  journal={arXiv preprint arXiv:\href{https://arxiv.org/abs/2510.01284}{2510.01284}},
  year={2025}
}

@article{huang2025jova,
  title={{JoVA}: Unified Multimodal Learning for Joint Video-Audio Generation},
  author={Huang, Xiaohu and Zhou, Hao and Yang, Qiangpeng and Wen, Shilei and Han, Kai},
  journal={arXiv preprint arXiv:\href{https://arxiv.org/abs/2512.13677}{2512.13677}},
  year={2025}
}

@article{blattmann2023stable,
  title={{Stable Video Diffusion}: Scaling latent video diffusion models to large datasets},
  author={Blattmann, Andreas and Dockhorn, Tim and Kulal, Sumith and Mendelevitch, Daniel and Kilian, Maciej and Lorenz, Dominik and Levi, Yam and English, Zion and Voleti, Vikram and Letts, Adam and others},
  journal={arXiv preprint arXiv:\href{https://arxiv.org/abs/2311.15127}{2311.15127}},
  year={2023}
}

@article{kong2024hunyuanvideo,
  title={{HunyuanVideo}: A systematic framework for large video generative models},
  author={Kong, Weijie and Tian, Qi and Zhang, Zijian and Min, Rox and Dai, Zuozhuo and Zhou, Jin and Xiong, Jiangfeng and Li, Xin and Wu, Bo and Zhang, Jianwei and others},
  journal={arXiv preprint arXiv:\href{https://arxiv.org/abs/2412.03603}{2412.03603}},
  year={2024}
}

@article{wan2025wan,
  title={{Wan}: Open and advanced large-scale video generative models},
  author={Wan, Team and Wang, Ang and Ai, Baole and Wen, Bin and Mao, Chaojie and Xie, Chen-Wei and Chen, Di and Yu, Feiwu and Zhao, Haiming and Yang, Jianxiao and others},
  journal={arXiv preprint arXiv:\href{https://arxiv.org/abs/2503.20314}{2503.20314}},
  year={2025}
}

@article{zhang2025waver,
  title={{Waver}: Wave your way to lifelike video generation},
  author={Zhang, Yifu and Yang, Hao and Zhang, Yuqi and Hu, Yifei and Zhu, Fengda and Lin, Chuang and Mei, Xiaofeng and Jiang, Yi and Peng, Bingyue and Yuan, Zehuan},
  journal={arXiv preprint arXiv:\href{https://arxiv.org/abs/2508.15761}{2508.15761}},
  year={2025}
}

@inproceedings{cheng2025mmaudio,
  title={{MMAudio}: Taming multimodal joint training for high-quality video-to-audio synthesis},
  author={Cheng, Ho Kei and Ishii, Masato and Hayakawa, Akio and Shibuya, Takashi and Schwing, Alexander and Mitsufuji, Yuki},
  booktitle={Proceedings of the Computer Vision and Pattern Recognition Conference},
  pages={28901--28911},
  year={2025}
}

@article{shan2025hunyuanvideo,
  title={{HunyuanVideo-Foley}: Multimodal diffusion with representation alignment for high-fidelity foley audio generation},
  author={Shan, Sizhe and Li, Qiulin and Cui, Yutao and Yang, Miles and Wang, Yuehai and Yang, Qun and Zhou, Jin and Zhong, Zhao},
  journal={arXiv preprint arXiv:\href{https://arxiv.org/abs/2508.16930}{2508.16930}},
  year={2025}
}

@article{wang2025kling,
  title={{Kling-Foley}: Multimodal diffusion transformer for high-quality video-to-audio generation},
  author={Wang, Jun and Zeng, Xijuan and Qiang, Chunyu and Chen, Ruilong and Wang, Shiyao and Wang, Le and Zhou, Wangjing and Cai, Pengfei and Zhao, Jiahui and Li, Nan and others},
  journal={arXiv preprint arXiv:\href{https://arxiv.org/abs/2506.19774}{2506.19774}},
  year={2025}
}

@article{chen2025hunyuanvideo,
  title={{HunyuanVideo-Avatar}: High-fidelity audio-driven human animation for multiple characters},
  author={Chen, Yi and Liang, Sen and Zhou, Zixiang and Huang, Ziyao and Ma, Yifeng and Tang, Junshu and Lin, Qin and Zhou, Yuan and Lu, Qinglin},
  journal={arXiv preprint arXiv:\href{https://arxiv.org/abs/2505.20156}{2505.20156}},
  year={2025}
}

@article{gan2025omniavatar,
  title={{OmniAvatar}: Efficient audio-driven avatar video generation with adaptive body animation},
  author={Gan, Qijun and Yang, Ruizi and Zhu, Jianke and Xue, Shaofei and Hoi, Steven},
  journal={arXiv preprint arXiv:\href{https://arxiv.org/abs/2506.18866}{2506.18866}},
  year={2025}
}

@article{gao2025wan,
  title={{Wan-S2V}: Audio-driven cinematic video generation},
  author={Gao, Xin and Hu, Li and Hu, Siqi and Huang, Mingyang and Ji, Chaonan and Meng, Dechao and Qi, Jinwei and Qiao, Penchong and Shen, Zhen and Song, Yafei and others},
  journal={arXiv preprint arXiv:\href{https://arxiv.org/abs/2508.18621}{2508.18621}},
  year={2025}
}

@misc{wan2.5,
  title={Wan2.5},
  author={Alibaba Cloud},
  note = {\url{https://wan.video/}},
  year={2025}
}

@misc{veo3,
  title={Veo3},
  author={Google DeepMind},
  note = {\url{https://deepmind.google/models/veo/}},
  year={2025}
}

@misc{sora2,
  title={Sora2},
  author={Openai},
  note = {\url{https://openai.com/index/sora-2/}},
  year={2025}
}

@article{hacohen2026ltx2efficientjointaudiovisual,
  title={{LTX-2}: Efficient Joint Audio-Visual Foundation Model},
  author={Yoav HaCohen and Benny Brazowski and Nisan Chiprut and Yaki Bitterman and Andrew Kvochko and Avishai Berkowitz and Daniel Shalem and Daphna Lifschitz and Dudu Moshe and Eitan Porat and Eitan Richardson and Guy Shiran and Itay Chachy and Jonathan Chetboun and Michael Finkelson and Michael Kupchick and Nir Zabari and Nitzan Guetta and Noa Kotler and Ofir Bibi and Ori Gordon and Poriya Panet and Roi Benita and Shahar Armon and Victor Kulikov and Yaron Inger and Yonatan Shiftan and Zeev Melumian and Zeev Farbman},
  journal={arXiv preprint arXiv:\href{https://arxiv.org/abs/2601.03233}{2601.03233}},
  year={2026}
}

@inproceedings{kang2024far,
  author       = {Bingyi Kang and
                  Yang Yue and
                  Rui Lu and
                  Zhijie Lin and
                  Yang Zhao and
                  Kaixin Wang and
                  Gao Huang and
                  Jiashi Feng},
  editor       = {Aarti Singh and
                  Maryam Fazel and
                  Daniel Hsu and
                  Simon Lacoste{-}Julien and
                  Felix Berkenkamp and
                  Tegan Maharaj and
                  Kiri Wagstaff and
                  Jerry Zhu},
  title        = {How Far Is Video Generation from World Model: {A} Physical Law Perspective},
  booktitle    = {International Conference on Machine Learning},
  series       = {Proceedings of Machine Learning Research},
  publisher    = {{PMLR} / OpenReview.net},
  year         = {2025},
  url          = {https://proceedings.mlr.press/v267/kang25g.html},
  bibsource    = {dblp computer science bibliography, https://dblp.org}
}

@article{deng2025denoising,
  author       = {Congyue Deng and
                  Brandon Y. Feng and
                  Cecilia Garraffo and
                  Alan Garbarz and
                  Robin Walters and
                  William T. Freeman and
                  Leonidas J. Guibas and
                  Kaiming He},
  title        = {Denoising {Hamiltonian} Network for Physical Reasoning},
  journal      = {Trans. Mach. Learn. Res.},
  volume       = {2026},
  year         = {2026},
  url          = {https://openreview.net/forum?id=KublEgx7Hv},
  bibsource    = {dblp computer science bibliography, https://dblp.org}
}

@inproceedings{chefer2025videojam,
  author       = {Hila Chefer and
                  Uriel Singer and
                  Amit Zohar and
                  Yuval Kirstain and
                  Adam Polyak and
                  Yaniv Taigman and
                  Lior Wolf and
                  Shelly Sheynin},
  editor       = {Aarti Singh and
                  Maryam Fazel and
                  Daniel Hsu and
                  Simon Lacoste{-}Julien and
                  Felix Berkenkamp and
                  Tegan Maharaj and
                  Kiri Wagstaff and
                  Jerry Zhu},
  title        = {{VideoJAM}: Joint Appearance-Motion Representations for Enhanced Motion
                  Generation in Video Models},
  booktitle    = {International Conference on Machine Learning},
  series       = {Proceedings of Machine Learning Research},
  publisher    = {{PMLR} / OpenReview.net},
  year         = {2025},
  url          = {https://proceedings.mlr.press/v267/chefer25a.html},
  bibsource    = {dblp computer science bibliography, https://dblp.org}
}

@inproceedings{liu2024physgen,
  title={{PhysGen}: Rigid-body physics-grounded image-to-video generation},
  author={Liu, Shaowei and Ren, Zhongzheng and Gupta, Saurabh and Wang, Shenlong},
  booktitle={European Conference on Computer Vision},
  pages={360--378},
  year={2024},
  organization={Springer}
}

@article{yuan2025newtongen,
  title={{NewtonGen}: Physics-Consistent and Controllable Text-to-Video Generation via Neural {Newtonian} Dynamics},
  author={Yuan, Yu and Wang, Xijun and Wickremasinghe, Tharindu and Nadir, Zeeshan and Ma, Bole and Chan, Stanley H},
  journal={arXiv preprint arXiv:\href{https://arxiv.org/abs/2509.21309}{2509.21309}},
  year={2025}
}

@inproceedings{zhang2025tora2,
  title={{Tora2}: Motion and appearance customized diffusion transformer for multi-entity video generation},
  author={Zhang, Zhenghao and Liao, Junchao and Meng, Xiangyu and Qin, Long and Wang, Weizhi},
  booktitle={Proceedings of the 33rd ACM International Conference on Multimedia},
  pages={9434--9443},
  year={2025}
}

@article{dai2023animateanythingfinegrainedopendomain,
  title={{AnimateAnything}: Fine-Grained Open Domain Image Animation with Motion Guidance},
  author={Zuozhuo Dai and Zhenghao Zhang and Yao Yao and Bingxue Qiu and Siyu Zhu and Long Qin and Weizhi Wang},
  journal={arXiv preprint arXiv:\href{https://arxiv.org/abs/2311.12886}{2311.12886}},
  year={2023}
}

@inproceedings{wang2024levitor,
  author       = {Hanlin Wang and
                  Hao Ouyang and
                  Qiuyu Wang and
                  Wen Wang and
                  Ka Leong Cheng and
                  Qifeng Chen and
                  Yujun Shen and
                  Limin Wang},
  title        = {{LeviTor}: {3D} Trajectory Oriented Image-to-Video Synthesis},
  booktitle    = {{IEEE/CVF} Conference on Computer Vision and Pattern Recognition},
  pages        = {12490--12500},
  publisher    = {Computer Vision Foundation / {IEEE}},
  year         = {2025},
  url          = {https://openaccess.thecvf.com/content/CVPR2025/html/Wang\_LeviTor\_3D\_Trajectory\_Oriented\_Image-to-Video\_Synthesis\_CVPR\_2025\_paper.html},
  doi          = {10.1109/CVPR52734.2025.01165},
  bibsource    = {dblp computer science bibliography, https://dblp.org}
}

@inproceedings{geng2025motionpromptingcontrollingvideo,
  author       = {Daniel Geng and
                  Charles Herrmann and
                  Junhwa Hur and
                  Forrester Cole and
                  Serena Zhang and
                  Tobias Pfaff and
                  Tatiana Lopez{-}Guevara and
                  Yusuf Aytar and
                  Michael Rubinstein and
                  Chen Sun and
                  Oliver Wang and
                  Andrew Owens and
                  Deqing Sun},
  title        = {Motion Prompting: Controlling Video Generation with Motion Trajectories},
  booktitle    = {{IEEE/CVF} Conference on Computer Vision and Pattern Recognition},
  pages        = {1--12},
  publisher    = {Computer Vision Foundation / {IEEE}},
  year         = {2025},
  url          = {https://openaccess.thecvf.com/content/CVPR2025/html/Geng\_Motion\_Prompting\_Controlling\_Video\_Generation\_with\_Motion\_Trajectories\_CVPR\_2025\_paper.html},
  doi          = {10.1109/CVPR52734.2025.00010},
  bibsource    = {dblp computer science bibliography, https://dblp.org}
}

@inproceedings{kingma2022autoencodingvariationalbayes,
  author       = {Diederik P. Kingma and
                  Max Welling},
  editor       = {Yoshua Bengio and
                  Yann LeCun},
  title        = {Auto-Encoding Variational {Bayes}},
  booktitle    = {International Conference on Learning Representations},
  year         = {2014},
  bibsource    = {dblp computer science bibliography, https://dblp.org}
}

@article{tjandra2025metaaudioboxaestheticsunified,
  title={{Meta Audiobox Aesthetics}: Unified Automatic Quality Assessment for Speech, Music, and Sound},
  author={Andros Tjandra and Yi-Chiao Wu and Baishan Guo and John Hoffman and Brian Ellis and Apoorv Vyas and Bowen Shi and Sanyuan Chen and Matt Le and Nick Zacharov and Carleigh Wood and Ann Lee and Wei-Ning Hsu},
  journal={arXiv preprint arXiv:\href{https://arxiv.org/abs/2502.05139}{2502.05139}},
  year={2025}
}

@inproceedings{gong2023contrastiveaudiovisualmaskedautoencoder,
  author       = {Yuan Gong and
                  Andrew Rouditchenko and
                  Alexander H. Liu and
                  David Harwath and
                  Leonid Karlinsky and
                  Hilde Kuehne and
                  James R. Glass},
  title        = {Contrastive Audio-Visual Masked Autoencoder},
  booktitle    = {International Conference on Learning Representations},
  publisher    = {OpenReview.net},
  year         = {2023},
  url          = {https://openreview.net/forum?id=QPtMRyk5rb},
  bibsource    = {dblp computer science bibliography, https://dblp.org}
}

@article{xie2025phyavbenchchallengingaudiophysicssensitivity,
  title={{PhyAVBench}: A Challenging Audio Physics-Sensitivity Benchmark for Physically Grounded Text-to-Audio-Video Generation},
  author={Tianxin Xie and Wentao Lei and Guanjie Huang and Pengfei Zhang and Kai Jiang and Chunhui Zhang and Fengji Ma and Haoyu He and Han Zhang and Jiangshan He and Jinting Wang and Linghan Fang and Lufei Gao and Orkesh Ablet and Peihua Zhang and Ruolin Hu and Shengyu Li and Weilin Lin and Xiaoyang Feng and Xinyue Yang and Yan Rong and Yanyun Wang and Zihang Shao and Zelin Zhao and Chenxing Li and Shan Yang and Wenfu Wang and Meng Yu and Dong Yu and Li Liu},
  journal={arXiv preprint arXiv:\href{https://arxiv.org/abs/2512.23994}{2512.23994}},
  year={2025}
}

@article{unterthiner2019accurategenerativemodelsvideo,
  title={Towards Accurate Generative Models of Video: A New Metric \& Challenges},
  author={Thomas Unterthiner and Sjoerd van Steenkiste and Karol Kurach and Raphael Marinier and Marcin Michalski and Sylvain Gelly},
  journal={arXiv preprint arXiv:\href{https://arxiv.org/abs/1812.01717}{1812.01717}},
  year={2019}
}

@inproceedings{wu2024largescalecontrastivelanguageaudiopretraining,
  author       = {Yusong Wu and
                  Ke Chen and
                  Tianyu Zhang and
                  Yuchen Hui and
                  Taylor Berg{-}Kirkpatrick and
                  Shlomo Dubnov},
  title        = {Large-Scale Contrastive Language-Audio Pretraining with Feature Fusion
                  and Keyword-to-Caption Augmentation},
  booktitle    = {{IEEE} International Conference on Acoustics, Speech and Signal Processing},
  pages        = {1--5},
  publisher    = {{IEEE}},
  year         = {2023},
  url          = {https://doi.org/10.1109/ICASSP49357.2023.10095969},
  doi          = {10.1109/ICASSP49357.2023.10095969},
  bibsource    = {dblp computer science bibliography, https://dblp.org}
}

@inproceedings{radford2021learningtransferablevisualmodels,
  author       = {Alec Radford and
                  Jong Wook Kim and
                  Chris Hallacy and
                  Aditya Ramesh and
                  Gabriel Goh and
                  Sandhini Agarwal and
                  Girish Sastry and
                  Amanda Askell and
                  Pamela Mishkin and
                  Jack Clark and
                  Gretchen Krueger and
                  Ilya Sutskever},
  editor       = {Marina Meila and
                  Tong Zhang},
  title        = {Learning Transferable Visual Models From Natural Language Supervision},
  booktitle    = {International Conference on Machine Learning},
  series       = {Proceedings of Machine Learning Research},
  pages        = {8748--8763},
  publisher    = {{PMLR}},
  year         = {2021},
  url          = {http://proceedings.mlr.press/v139/radford21a.html},
  bibsource    = {dblp computer science bibliography, https://dblp.org}
}

@inproceedings{chen2020vggsoundlargescaleaudiovisualdataset,
  author       = {Honglie Chen and
                  Weidi Xie and
                  Andrea Vedaldi and
                  Andrew Zisserman},
  title        = {{VGGSound}: {A} Large-Scale Audio-Visual Dataset},
  booktitle    = {{IEEE} International Conference on Acoustics, Speech and Signal Processing},
  pages        = {721--725},
  publisher    = {{IEEE}},
  year         = {2020},
  url          = {https://doi.org/10.1109/ICASSP40776.2020.9053174},
  doi          = {10.1109/ICASSP40776.2020.9053174},
  bibsource    = {dblp computer science bibliography, https://dblp.org}
}

@inproceedings{lee2021acav100mautomaticcurationlargescale,
  author       = {Sangho Lee and
                  Jiwan Chung and
                  Youngjae Yu and
                  Gunhee Kim and
                  Thomas M. Breuel and
                  Gal Chechik and
                  Yale Song},
  title        = {{ACAV100M:} Automatic Curation of Large-Scale Datasets for Audio-Visual
                  Video Representation Learning},
  booktitle    = {{IEEE/CVF} International Conference on Computer Vision},
  pages        = {10254--10264},
  publisher    = {{IEEE}},
  year         = {2021},
  url          = {https://doi.org/10.1109/ICCV48922.2021.01011},
  doi          = {10.1109/ICCV48922.2021.01011},
  bibsource    = {dblp computer science bibliography, https://dblp.org}
}

@inproceedings{nan2025openvid1mlargescalehighqualitydataset,
  author       = {Kepan Nan and
                  Rui Xie and
                  Penghao Zhou and
                  Tiehan Fan and
                  Zhenheng Yang and
                  Zhijie Chen and
                  Xiang Li and
                  Jian Yang and
                  Ying Tai},
  title        = {{OpenVid-1M}: {A} Large-Scale High-Quality Dataset for Text-to-video
                  Generation},
  booktitle    = {International Conference on Learning Representations},
  publisher    = {OpenReview.net},
  year         = {2025},
  url          = {https://openreview.net/forum?id=j7kdXSrISM},
  bibsource    = {dblp computer science bibliography, https://dblp.org}
}

@misc{pexel,
  author = {{Pexels}},
  title = {Pexels},
  note = {\url{https://www.pexels.com/}},
}

@article{bai2025qwen3vltechnicalreport,
  title={{Qwen3-VL} Technical Report},
  author={Shuai Bai and Yuxuan Cai and Ruizhe Chen and Keqin Chen and Xionghui Chen and Zesen Cheng and Lianghao Deng and Wei Ding and Chang Gao and Chunjiang Ge and Wenbin Ge and Zhifang Guo and Qidong Huang and Jie Huang and Fei Huang and Binyuan Hui and Shutong Jiang and Zhaohai Li and Mingsheng Li and Mei Li and Kaixin Li and Zicheng Lin and Junyang Lin and Xuejing Liu and Jiawei Liu and Chenglong Liu and Yang Liu and Dayiheng Liu and Shixuan Liu and Dunjie Lu and Ruilin Luo and Chenxu Lv and Rui Men and Lingchen Meng and Xuancheng Ren and Xingzhang Ren and Sibo Song and Yuchong Sun and Jun Tang and Jianhong Tu and Jianqiang Wan and Peng Wang and Pengfei Wang and Qiuyue Wang and Yuxuan Wang and Tianbao Xie and Yiheng Xu and Haiyang Xu and Jin Xu and Zhibo Yang and Mingkun Yang and Jianxin Yang and An Yang and Bowen Yu and Fei Zhang and Hang Zhang and Xi Zhang and Bo Zheng and Humen Zhong and Jingren Zhou and Fan Zhou and Jing Zhou and Yuanzhi Zhu and Ke Zhu},
  journal={arXiv preprint arXiv:\href{https://arxiv.org/abs/2511.21631}{2511.21631}},
  year={2025}
}

@article{xu2025qwen3omnitechnicalreport,
  title={{Qwen3-Omni} Technical Report},
  author={Jin Xu and Zhifang Guo and Hangrui Hu and Yunfei Chu and Xiong Wang and Jinzheng He and Yuxuan Wang and Xian Shi and Ting He and Xinfa Zhu and Yuanjun Lv and Yongqi Wang and Dake Guo and He Wang and Linhan Ma and Pei Zhang and Xinyu Zhang and Hongkun Hao and Zishan Guo and Baosong Yang and Bin Zhang and Ziyang Ma and Xipin Wei and Shuai Bai and Keqin Chen and Xuejing Liu and Peng Wang and Mingkun Yang and Dayiheng Liu and Xingzhang Ren and Bo Zheng and Rui Men and Fan Zhou and Bowen Yu and Jianxin Yang and Le Yu and Jingren Zhou and Junyang Lin},
  journal={arXiv preprint arXiv:\href{https://arxiv.org/abs/2509.17765}{2509.17765}},
  year={2025}
}

@inproceedings{ravi2024sam2segmentimages,
  author       = {Nikhila Ravi and
                  Valentin Gabeur and
                  Yuan{-}Ting Hu and
                  Ronghang Hu and
                  Chaitanya Ryali and
                  Tengyu Ma and
                  Haitham Khedr and
                  Roman R{\"{a}}dle and
                  Chlo{\'{e}} Rolland and
                  Laura Gustafson and
                  Eric Mintun and
                  Junting Pan and
                  Kalyan Vasudev Alwala and
                  Nicolas Carion and
                  Chao{-}Yuan Wu and
                  Ross B. Girshick and
                  Piotr Doll{\'{a}}r and
                  Christoph Feichtenhofer},
  title        = {{SAM} 2: Segment Anything in Images and Videos},
  booktitle    = {International Conference on Learning Representations},
  publisher    = {OpenReview.net},
  year         = {2025},
  url          = {https://openreview.net/forum?id=Ha6RTeWMd0},
  bibsource    = {dblp computer science bibliography, https://dblp.org}
}

@inproceedings{yang2022maniqamultidimensionattentionnetwork,
  author       = {Sidi Yang and
                  Tianhe Wu and
                  Shuwei Shi and
                  Shanshan Lao and
                  Yuan Gong and
                  Mingdeng Cao and
                  Jiahao Wang and
                  Yujiu Yang},
  title        = {{MANIQA:} Multi-dimension Attention Network for No-Reference Image
                  Quality Assessment},
  booktitle    = {{IEEE/CVF} Conference on Computer Vision and Pattern Recognition Workshops},
  pages        = {1190--1199},
  publisher    = {{IEEE}},
  year         = {2022},
  url          = {https://doi.org/10.1109/CVPRW56347.2022.00126},
  doi          = {10.1109/CVPRW56347.2022.00126},
  bibsource    = {dblp computer science bibliography, https://dblp.org}
}

@inproceedings{ke2021musiqmultiscaleimagequality,
  author       = {Junjie Ke and
                  Qifei Wang and
                  Yilin Wang and
                  Peyman Milanfar and
                  Feng Yang},
  title        = {{MUSIQ:} Multi-scale Image Quality Transformer},
  booktitle    = {{IEEE/CVF} International Conference on Computer Vision},
  pages        = {5128--5137},
  publisher    = {{IEEE}},
  year         = {2021},
  url          = {https://doi.org/10.1109/ICCV48922.2021.00510},
  doi          = {10.1109/ICCV48922.2021.00510},
  bibsource    = {dblp computer science bibliography, https://dblp.org}
}

@article{openmossteam2026movascalablesynchronizedvideoaudio,
  title={{MOVA}: Towards Scalable and Synchronized Video-Audio Generation},
  author={OpenMOSS Team and Donghua Yu and Mingshu Chen and Qi Chen and Qi Luo and Qianyi Wu and Qinyuan Cheng and Ruixiao Li and Tianyi Liang and Wenbo Zhang and Wenming Tu and Xiangyu Peng and Yang Gao and Yanru Huo and Ying Zhu and Yinze Luo and Yiyang Zhang and Yuerong Song and Zhe Xu and Zhiyu Zhang and Chenchen Yang and Cheng Chang and Chushu Zhou and Hanfu Chen and Hongnan Ma and Jiaxi Li and Jingqi Tong and Junxi Liu and Ke Chen and Shimin Li and Shiqi Jiang and Songlin Wang and Wei Jiang and Zhaoye Fei and Zhiyuan Ning and Chunguo Li and Chenhui Li and Ziwei He and Zengfeng Huang and Xie Chen and Xipeng Qiu},
  journal={arXiv preprint arXiv:\href{https://arxiv.org/abs/2602.08794}{2602.08794}},
  year={2026}
}

@article{DBLP:journals/ijon/SuALPBL24,
  author       = {Jianlin Su and
                  Murtadha H. M. Ahmed and
                  Yu Lu and
                  Shengfeng Pan and
                  Wen Bo and
                  Yunfeng Liu},
  title        = {{RoFormer}: Enhanced transformer with Rotary Position Embedding},
  journal      = {Neurocomputing},
  volume       = {568},
  pages        = {127063},
  year         = {2024},
  url          = {https://doi.org/10.1016/j.neucom.2023.127063},
  doi          = {10.1016/J.NEUCOM.2023.127063},
  bibsource    = {dblp computer science bibliography, https://dblp.org}
}

@article{benyosef2026avcontrolefficientframeworktraining,
  title={{AVControl}: Efficient Framework for Training Audio-Visual Controls},
  author={Matan Ben-Yosef and Tavi Halperin and Naomi Ken Korem and Mohammad Salama and Harel Cain and Asaf Joseph and Anthony Chen and Urska Jelercic and Ofir Bibi},
  journal={arXiv preprint arXiv:\href{https://arxiv.org/abs/2603.24793}{2603.24793}},
  year={2026}
}

@inproceedings{DBLP:conf/cvpr/RombachBLEO22,
  author       = {Robin Rombach and
                  Andreas Blattmann and
                  Dominik Lorenz and
                  Patrick Esser and
                  Bj{\"{o}}rn Ommer},
  title        = {High-Resolution Image Synthesis with Latent Diffusion Models},
  booktitle    = {{IEEE/CVF} Conference on Computer Vision and Pattern Recognition},
  pages        = {10674--10685},
  publisher    = {{IEEE}},
  year         = {2022},
  url          = {https://doi.org/10.1109/CVPR52688.2022.01042},
  doi          = {10.1109/CVPR52688.2022.01042},
  bibsource    = {dblp computer science bibliography, https://dblp.org}
}

@inproceedings{DBLP:conf/eccv/WuLGZHZSLGZ24,
  author       = {Weijia Wu and
                  Zhuang Li and
                  Yuchao Gu and
                  Rui Zhao and
                  Yefei He and
                  David Junhao Zhang and
                  Mike Zheng Shou and
                  Yan Li and
                  Tingting Gao and
                  Di Zhang},
  editor       = {Ales Leonardis and
                  Elisa Ricci and
                  Stefan Roth and
                  Olga Russakovsky and
                  Torsten Sattler and
                  G{\"{u}}l Varol},
  title        = {{DragAnything}: Motion Control for Anything Using Entity Representation},
  booktitle    = {European Conference on Computer Vision},
  series       = {Lecture Notes in Computer Science},
  pages        = {331--348},
  publisher    = {Springer},
  year         = {2024},
  url          = {https://doi.org/10.1007/978-3-031-72670-5\_19},
  doi          = {10.1007/978-3-031-72670-5\_19},
  bibsource    = {dblp computer science bibliography, https://dblp.org}
}

@inproceedings{karaev2024cotracker3simplerbetterpoint, title={{CoTracker3}: Simpler and Better Point Tracking by Pseudo-Labeling Real Videos}, url={http://dx.doi.org/10.1109/iccv51701.2025.00568}, DOI={10.1109/iccv51701.2025.00568}, booktitle={2025 IEEE/CVF International Conference on Computer Vision (ICCV)}, publisher={IEEE}, author={Karaev, Nikita and Makarov, Yuri and Wang, Jianyuan and Neverova, Natalia and Vedaldi, Andrea and Rupprecht, Christian}, year={2025}, month=Oct, pages={1–10} }

@article{qiang2025transferability,
  title={On the Transferability and Discriminability of Representation Learning in Unsupervised Domain Adaptation},
  author={Qiang, Wenwen and Gu, Ziyin and Si, Lingyu and Li, Jiangmeng and Zheng, Changwen and Sun, Fuchun and Xiong, Hui},
  journal={IEEE Transactions on Pattern Analysis and Machine Intelligence},
  year={2025},
  publisher={IEEE}
}

@inproceedings{zhang2026enhancing,
  title={Enhancing Large Language Models for Time-Series Forecasting via Vector-Injected In-Context Learning},
  author={Zhang, Jianqi and Wang, Jingyao and Qiang, Wenwen and Xu, Fanjiang and Zheng, Changwen},
  booktitle={Proceedings of the ACM Web Conference 2026},
  pages={7211--7222},
  year={2026}
}

@article{qiang2026plasticity,
  title={On the Plasticity and Stability for Post-Training Large Language Models},
  author={Qiang, Wenwen and Gu, Ziyin and Zhou, Jiahuan and Hu, Jie and Wang, Jingyao and Zheng, Changwen and Xiong, Hui},
  journal={arXiv preprint arXiv:\href{https://arxiv.org/abs/2602.06453}{2602.06453}},
  year={2026}
}

\end{document}